\documentclass[twoside]{article}

\usepackage[accepted]{aistats2020}
\usepackage[round]{natbib}

\usepackage{amsfonts}       
\usepackage{nicefrac}       
\usepackage{microtype}      

\usepackage{hyperref}

\usepackage{arydshln}
\usepackage{booktabs}

\usepackage{amsmath}
\usepackage{amssymb}
\usepackage{algorithm2e}
\usepackage{subfigure}
\usepackage{graphicx}
\usepackage{multirow}
\usepackage{cite}
\usepackage{color,soul}

\begin{document}

%

%

\twocolumn[

\aistatstitle{Learnable Bernoulli Dropout for Bayesian Deep Learning}

\aistatsauthor{ Shahin Boluki$^{\dagger}$ \And Randy Ardywibowo$^{\dagger}$ \And  Siamak Zamani Dadaneh$^{\dagger}$  }

\aistatsauthor{ Mingyuan Zhou$^{\star}$ \And Xiaoning Qian$^{\dagger}$}

\aistatsaddress{ $^{\dagger}$ Texas A\&M University \And  $^{\star}$The University of Texas at Austin } ]

\begin{abstract}
In this work, we propose learnable Bernoulli dropout (LBD), a new model-agnostic dropout scheme that considers the dropout rates as parameters jointly optimized with other model parameters. By probabilistic modeling of Bernoulli dropout, our method enables more robust prediction and uncertainty quantification in deep models. Especially, when combined with variational auto-encoders (VAEs), LBD enables flexible semi-implicit posterior representations, leading to new semi-implicit VAE~(SIVAE) models. We solve the optimization for training with respect to the dropout parameters using Augment-REINFORCE-Merge (ARM), an unbiased and low-variance gradient estimator. Our experiments on a range of tasks show the superior performance of our approach compared with other commonly used dropout schemes. Overall, LBD leads to improved accuracy and uncertainty estimates in image classification and semantic segmentation. Moreover, using SIVAE, we can achieve state-of-the-art performance on collaborative filtering for implicit feedback on several public datasets.

\end{abstract}
\section{\uppercase{Introduction}}

Deep neural networks (DNNs) are a flexible family of models that usually contain millions of free parameters. Growing concerns on overfitting of DNNs \citep{szegedy2013intriguing,nguyen2015deep,zhang2016understanding,bozhinoski2019safety} arise especially when considering their robustness and generalizability in real-world safety-critical applications such as autonomous driving and healthcare \citep{ardywibowo2019adaptive}. To address this, Bayesian methods attempt to principally regularize and estimate the prediction uncertainty of DNNs. They introduce model uncertainty by placing prior distributions on the weights and biases of the networks. Since exact Bayesian inference is computationally intractable, many approximation methods have been developed such as Laplace approximation \citep{mackay1992bayesian}, Markov chain Monte Carlo~(MCMC) \citep{neal2012bayesian}, stochastic gradient MCMC \citep{welling2011bayesian,ma2015complete,springenberg2016bayesian}, and variational inference methods \citep{blei2017variational,hoffman2013stochastic,blundell2015weight, graves2011practical}. In practice, these methods are significantly slower to train compared to non-Bayesian methods for DNNs, such as calibrated regression \citep{kuleshov2018accurate}, deep ensemble methods \citep{lakshminarayanan2017simple}, and more recent prior networks \citep{malinin2018predictive}, which have their own limitations including training instability \citep{blum2019fishyscapes}.

Although dropout, a commonly used technique to alleviate overfitting in DNNs, was initially used as a regularization technique during training~\citep{hinton2012improving}, \citet{gal2016dropout} showed that when used at test time, it enables uncertainty quantification with Bayesian interpretation of the network outputs as Monte Carlo samples of its predictive distribution. Considering the original dropout scheme as multiplying the output of each neuron by a binary mask drawn from a Bernoulli distribution, several dropout variations with other distributions for random multiplicative masks have been studied, including Gaussian dropout \citep{kingma2015variational,srivastava2014dropout}. Among them, Bernoulli dropout and extensions are most commonly used in practice due to their ease of implementation in existing deep architectures and their computation speed. Its simplicity and computational tractability has made Bernoulli dropout the current most popular method to introduce uncertainty in DNNs.

It has been shown that both the level of prediction accuracy and quality of uncertainty estimation are dependent on the network weight configuration as well as the dropout rate \citep{gal2016uncertainty}. Traditional dropout mechanism with fixed dropout rate may limit model expressiveness or require tedious hand-tuning. Allowing the dropout rate to be estimated along with the other network parameters increases model flexibility and enables feature sparsity patterns to be identified. An early approach to learning dropout rates overlays a binary belief network on top of DNNs to determine dropout rates \citep{ba2013adaptive}. Unfortunately, this approach does not scale well due to the significant model complexity increase. 

Other dropout formulations instead attempt to replace the Bernoulli dropout with a different distribution. Following the variational interpretation of Gaussian dropout, \citet{kingma2015variational} proposed to optimize the variance of the Gaussian distributions used for the multiplicative masks. However, in practice, optimization of the Gaussian variance is difficult. For example, the variance should be capped at 1 in order to prevent the optimization from diverging. This assumption limits the dropout rate to at most 0.5, and is not suitable for regularizing architectures with potentially redundant features, which should be dropped at higher rates. Also, \citet{hron2017variational} showed that approximate Bayesian inference of Gaussian dropout is ill-posed since the improper log-uniform prior adopted in  \citep{kingma2015variational} does not usually result in a proper posterior. Recently, a relaxed Concrete \citep{maddison2016concrete} (Gumbell-Softmax \citep{jang2016categorical}) distribution has been adopted in \citep{gal2017concrete} to replace the Bernoulli mask for learnable dropout rate \citep{gal2016uncertainty}. However, the continuous relaxation introduces bias to the gradients which reduces its performance.

Motivated by recent efforts on gradient estimates for optimization with binary (discrete) variables~\citep{yin2018arm,tucker2017rebar,grathwohl2017backpropagation}, we propose a learnable Bernoulli dropout (LBD) module for general DNNs. In LBD, the dropout probabilities are considered as variational parameters jointly optimized with the other parameters of the model. We emphasize that LBD exactly optimizes the true Bernoulli distribution of regular dropout, instead of replacing it by another distribution such as Concrete or Gaussian. LBD accomplishes this by taking advantage of a recent unbiased low-variance gradient estimator---Augment-REINFORCE-Merge (ARM) \citep{yin2018arm}---to optimize the loss function of the deep neural network with respect to the dropout layer. This allows us to backpropagate through the binary random masks and compute unbiased, low-variance gradients with respect to the dropout parameters. This approach properly introduces learnable feature sparsity regularization to the deep network, improving the performance of deep architectures that rely on it. Moreover, our formulation allows each neuron to have its own learned dropout probability. We provide an interpretation of LBD as a more flexible variational Bayesian approximation method for learning Bayesian DNNs compared to Monte Carlo (MC) dropout. We combine this learnable dropout module with variational autoencoders (VAEs) \citep{kingma2013auto,rezende2014stochastic}, which naturally leads to a flexible semi-implicit variational inference framework with VAEs (SIVAE). Our experiments show that LBD results in improved accuracy and uncertainty quantification in DNNs for image classification and semantic segmentation compared with regular dropout, MC dropout \citep{gal2016dropout}, Gaussian dropout \citep{kingma2015variational}, and Concrete dropout \citep{gal2017concrete}. More importantly, by performing optimization of the dropout rates in SIVAE, we achieve state-of-the-art performance in multiple different collaborative filtering benchmarks.

\section{\uppercase  {Methodology}}

\subsection{Learnable Bernoulli Dropout (LBD)}\label{sec:dropout}

Given a training dataset $\mathcal{D}=\{(\boldsymbol{x}_i,y_i)\}_{i=1}^{N}$, where $\boldsymbol{x}$ and $y$ denote the input and target of interest respectively, a neural network is a function $f(\boldsymbol{x};\boldsymbol{\theta})$ from the input space to the target space with parameters $\boldsymbol{\theta}$. The parameters are  learned by minimizing an objective function $\mathcal{L}$, which is usually comprised of an empirical loss $\mathcal{E}$ with possibly additional regularization terms $\mathcal{R}$, by stochastic gradient descent (SGD):
\begin{equation} \label{eq:gen}
\mathcal{L}(\boldsymbol{\theta} | \mathcal{D}) \approx \frac{N}{M} \sum_{i=1}^{M}\mathcal{E}(f(\boldsymbol{x}_i;\boldsymbol{\theta}),y_i) + \mathcal{R(\boldsymbol{\theta})},
\end{equation}
where $M$ is the mini-batch size.


Consider a neural network with $L$ fully connected layers. The $j^{th}$ fully connected layer with $K_j$ neurons takes the output of the $(j-1)^{th}$ layer with $K_{j-1}$ neurons as input. We denote the weight matrix connecting layer $j-1$ to $j$ by $W_j \in \mathbb{R}^{K_{j-1}\times K_{j}}$. Dropout takes the output to each layer and multiplies it with a random variable $\boldsymbol{z}_j \sim p(\boldsymbol{z}_j)$ element-wise (channel-wise for convolutional layers). The most common choice for $p(\boldsymbol{z}_j)$ is the Bernoulli distribution $\text{Ber}(\sigma(\alpha_j))$ with dropout rate $1-\sigma(\alpha_j)$, where we have reparameterized the dropout rate using the sigmoid function $\sigma(\cdot)$. With this notation, let $\boldsymbol{\alpha}=\{\alpha_j\}_{j=1}^{L}$ denote the collection of all logits of the dropout parameters, and let $\boldsymbol{z}=\{\boldsymbol{z}_j\}_{j=1}^{L}$ denote the collection of all dropout masks. Dropout in this form is one of the most common regularization techniques in training DNNs to avoid overfitting and improve generalization and accuracy on unseen data. This can also be considered as using a data dependent weight; if $z_{jk}=0$ for input $\boldsymbol{x}$, then the $k$th row of $W_j$ will be set to zero.

The parameter $\boldsymbol{\alpha}$ of the random masks $\boldsymbol{z}$ has been mainly treated as hyperparameters in the literature, requiring tuning by grid-search, which is prohibitively expensive. Instead, we propose to learn the dropout rates for Bernoulli masks jointly with the other model parameters. Specifically, we aim to optimize the expectation of the loss function with respect to the Bernoulli distributed dropouts:
\begin{equation}
   \min_{\boldsymbol{\theta}=\{\boldsymbol{\theta}\setminus \boldsymbol{\alpha},\boldsymbol{\alpha}\}} ~~ \mathbb{E}_{\boldsymbol{z}\sim\prod_{i=1}^M \text{Ber}(\boldsymbol{z}_i;\sigma(\boldsymbol{\alpha}))} \big[\mathcal{L}(\boldsymbol{\theta},\boldsymbol{z} | \mathcal{D}) \big].
   \label{eq:obj}
\end{equation}

We next formulate the problem of learning dropout rates for supervised feed-forward DNNs, and unsupervised latent representation learning in VAEs. In the formulations that follow, the dropout rates can be optimized using the method described in Section \ref{sec:optimization}. We first briefly review the variational interpretation of dropout in Bayesian deep neural networks and show how our LBD fits here. Then, we discuss how the dropout rates can be adaptive to the data in the context of VAEs. Specifically, combining the Bernoulli dropout layer with VAEs allows us to construct a semi-implicit variational inference framework.

\subsection{Variational Bayesian Inference with LBD}
\label{sec:Bayesian}
In Bayesian neural networks (BNNs) \citep{mackay1992practical,neal95}, instead of finding point estimates of the weight matrices, the goal is to learn a distribution over them. In this setup, a prior is assumed over the weight matrices, $p(W)$, which is updated to a posterior given the training data following Bayes' rule $p(W | \mathcal{D})=\frac{p(\mathcal{D} | W)p(W)}{p(\mathcal{D})}$. This posterior distribution captures the set of plausible models and imposes a predictive distribution for the target of a new data point. Due to the intractability of calculating $p(\mathcal{D})$, different approximation techniques have been developed \citep{blei2017variational,graves2011practical,blundell2015weight,gal2016dropout}, which use a simple (variational) distribution $q_{\boldsymbol{\theta}}(W)$ to approximate the posterior. By taking this approach the intractable marginalization in the original inference problem is replaced by an optimization problem, where the parameters $\boldsymbol{\theta}$ are optimized by fitting $q_{\boldsymbol{\theta}}(W)$ to $p(W | \mathcal{D})$. Following the variational interpretation of Bernoulli dropout, the approximating distribution is a mixture of two Gaussian distributions with very small variance, where one mixture component has mean zero \citep{gal2016dropout,gal2016bayesian}. Assuming a network with $L$ layers, we denote the collection of all weight matrices by $\boldsymbol{W}=\{W_j\}_{j=1}^{L}$.

Under this formulation, we propose learnable Bernoulli dropout~(LBD) as a variational approximation. Unlike the common assumption where all neurons in layer $j$ share the same dropout rate, in our approach, we let each neuron $k$ in each layer have its own dropout rate $\alpha_{jk}$. Thus, each layer has a mean weight matrix $M_j$ and dropout parameters $\boldsymbol{\alpha}_{j}=\{\alpha_{jk}\}_{k=1}^{K_{j-1}}$. With this, our variational distribution consists of the parameters $\boldsymbol{\theta}=\{M_j,\boldsymbol{\alpha}_{j}\}_{j=1}^{L}$. In this setup, 
$q_{\boldsymbol{\theta}}(\boldsymbol{W}) = \prod_{j=1}^{L}q_{\boldsymbol{\theta}}(W_j)$, where $q_{\boldsymbol{\theta}}(W_j) = M_j^T\text{diag}(\text{Ber}(\boldsymbol{\alpha}_{j}))$, and the objective function for optimizing the variational parameters is  
\begin{equation} 
\begin{split}
  \small\mathcal{L}(\boldsymbol{\theta}=\{M_j,\boldsymbol{\alpha}_{j}\}_{j=1}^{L}  |  \mathcal{D})&=-\frac{N}{M} \sum_{i=1}^{M} \log p(y_i | f(\boldsymbol{x}_i;\boldsymbol{W}_i)) \\&+ \text{KL}(q_{\boldsymbol{\theta}}(\boldsymbol{W})||p(\boldsymbol{W})), 
\end{split}\label{eq:Bayesian}
\end{equation}
where $\boldsymbol{W}_i$ denotes the realization of the random weight matrices for each data point. The likelihood function $p(y_i | f(\boldsymbol{x}_i;\boldsymbol{W}_i))$ is usually a softmax or a Gaussian for classification and regression problems, respectively. The Kullback-Leibler~(KL) divergence term is a regularization term that prevents the approximate posterior from deviating too far from the prior. By employing the quantized zero-mean Gaussian prior in \citep{gal2016uncertainty} with variance $s^2$, we have $\text{KL}(q_{\boldsymbol{\theta}}(\boldsymbol{W})||p(\boldsymbol{W})) \propto \sum_{j=1}^{L} \sum_{k=1}^{K_{j-1}} \frac{\alpha_{jk}}{2s^2}||M_j[\cdot,k]||^2 -\mathcal{H}(\alpha_{jk})$, where $M_j[\cdot,k]$ represents the $k^{\text{th}}$ column of the mean weight matrix $M_j$ and $\mathcal{H}(\alpha_{jk})$ denotes the entropy of a Bernoulli random variable with parameter $\alpha_{jk}$. 

After fitting the approximate posterior distribution, the posterior predictive $p(y | \boldsymbol{x},\mathcal{D})$ for the target of a new data point $\boldsymbol{x}$ is approximated by Monte Carlo integration with $S$ samples obtained by stochastic forward passes as $\frac{1}{S}\sum_{s=1}^{S}p(y | f(\boldsymbol{x};\boldsymbol{W}^{(s)})$. The entropy of the posterior predictive $p(y | \boldsymbol{x},\mathcal{D})$ can be considered as a measure of uncertainty \citep{mukhoti2018evaluating,gal2016uncertainty}.

Note that the variational interpretation and the derivation of the objective function in this section corresponds to Bernoulli dropout, i.e. when the variational distributions are constructed by multiplications with Bernoulli random variables. Replacing the Bernoulli distribution with a relaxed distribution like Concrete introduces an additional level of approximation and bias which can lead to degraded predictive and uncertainty quantification performance.

\subsection{Combining VAE and LBD into SIVAE}
\label{sec:SIVAE}

VAEs \citep{kingma2013auto,rezende2014stochastic} have been widely used for amortized inference and unsupervised feature learning. They tie together the modeling and inference through an encoder-decoder architecture. The encoder (data-dependent variational posterior) $q_{\boldsymbol{\psi}}(\boldsymbol{\eta} | \boldsymbol{x}_i)$ and decoder
$p_{\boldsymbol{\omega}}(\boldsymbol{x}_i | \boldsymbol{\eta})$ (generative model) are based on neural networks parameterized by $\boldsymbol{\psi}$ and $\boldsymbol{\omega}$, respectively, which are inferred by minimizing the negative evidence lower bound (ELBO):
\begin{equation}\label{eq:VAE}
\begin{split}
    \mathcal{L}(\boldsymbol{\theta}=\{\boldsymbol{\psi},\boldsymbol{\omega}\} | \mathcal{D}) &= -\frac{1}{N} \sum_{i=1}^N\mathbb{E}_{q_{\boldsymbol{\psi}}(\boldsymbol{\eta} | \boldsymbol{x}_i)}\big[\log p_{\boldsymbol{\omega}}(\boldsymbol{x}_i | \boldsymbol{\eta})\big] \\&+ \quad \beta \text{KL}(q_{\boldsymbol{\psi}}(\boldsymbol{\eta} | \boldsymbol{x}_i)||p(\boldsymbol{\eta})),
\end{split}
\end{equation}
where the prior over the latent variables, $p(\boldsymbol{\eta})$, is commonly set to $\mathcal{N}(0,I)$. While $\beta=1$ in the original VAE \citep{kingma2013auto}, setting either $\beta>1$ or $\beta<1$ have been proposed \citep{higgins2016beta,zhao2018information} to allow for learning more informative latent representations or maximization of the mutual information between latent and observed variables. Regular dropout has been commonly used in training VAEs to prevent overfitting. When training, one sample is taken for each observed point in the mini-batch and the network parameters are found by minimizing~\eqref{eq:VAE}. The dropout rate can be tuned by grid-search to find the highest ELBO. This approach is prohibitively expensive for large (deep/wide) models. 

Instead of using Gaussian for the data-dependent variational distribution $q_{\boldsymbol{\psi}}(\boldsymbol{\eta} | \boldsymbol{x}) = \mathcal{N}(\mu_{\boldsymbol{\psi}}(\boldsymbol{x}),\text{diag}(\Sigma_{\boldsymbol{\psi}}(\boldsymbol{x})))$, we adopt the semi-implicit variational inference (SIVI) approach \citep{yin2018semi}, allowing us to use more expressive data-dependent variational distributions. Here, the variational distribution consists of an explicit conditional distribution $q(\boldsymbol{\eta} | \boldsymbol{\gamma})$ mixed with an implicit distribution $q(\boldsymbol{\gamma})$. This allows the construction of a semi-implicit VAE (SIVAE), where the first stochastic layer, $q_{\boldsymbol{\psi}}(\boldsymbol{\eta} | \boldsymbol{x})$, is still a Gaussian distribution, but other stochastic layers are constructed implicitly in the upper layers of the encoder. In \citep{yin2018semi}, the stochastic layers were constructed by concatenating the deterministic layer output with random noise and observed data to form the input to the next layer.

In our approach, dropout can be considered to introduce an implicit mixing effect that constructs a SIVAE. More specifically, a realization of the local dropout mask results in a predicted mean and variance for the Gaussian variational distribution $\mathcal{N}(\mu_{\boldsymbol{\psi}}(\boldsymbol{x},\boldsymbol{z}),\text{diag}(\Sigma_{\boldsymbol{\psi}}(\boldsymbol{x}
,\boldsymbol{z})))$. By marginalizing over the dropout, we can construct the following implicit variational distribution:
\begin{equation}
\begin{split}
q_{\boldsymbol{\psi}}(\boldsymbol{\eta} | \boldsymbol{x})=&\mathbb{E}_{\boldsymbol{z}\sim p_{\boldsymbol{\alpha}}(\boldsymbol{z}) }[q_{\boldsymbol{\psi}}(\boldsymbol{\eta} | \boldsymbol{x},\boldsymbol{z})]\\=&\mathbb{E}_{\boldsymbol{z}\sim p_{\boldsymbol{\alpha}}(\boldsymbol{z}) }[\mathcal{N}(\mu_{\boldsymbol{\psi}}(\boldsymbol{x},\boldsymbol{z}),\text{diag}(\Sigma_{\boldsymbol{\psi}}(\boldsymbol{x}
,\boldsymbol{z})))],
\end{split}
\end{equation}
which is more flexible than the common Gaussian assumption and can result in better latent representations. Using an asymptotically exact ELBO for SIVAE \citep{yin2018semi}, we infer both the encoder and decoder network parameters together with the dropout rates by optimizing
\begin{equation}
\begin{split}
    &\mathcal{L}(\boldsymbol{\theta}=\{\boldsymbol{\psi},\boldsymbol{\omega},\boldsymbol{\alpha}\} | \mathcal{D}) =\\& -\frac{1}{N} \sum_{i=1}^N \mathbb{E}_{\boldsymbol{z}_i\sim p_{\boldsymbol{\alpha}}(\boldsymbol{z}), \boldsymbol{\eta} \sim q_{\boldsymbol{\psi}}(\boldsymbol{\eta} | \boldsymbol{x}_i,\boldsymbol{z}_i), \boldsymbol{z}_i^{(1)},\cdots, \boldsymbol{z}_i^{(V)}\sim  p_{\boldsymbol{\alpha}}(\boldsymbol{z}) }\Big[ \\& \log p_{\boldsymbol{\omega}}(\boldsymbol{x}_i | \boldsymbol{\eta})p(\boldsymbol{\eta})-\log \frac{1}{V+1} \big[q_{\boldsymbol{\psi}}(\boldsymbol{\eta} | \boldsymbol{x}_i,\boldsymbol{z}_i) +\\&\quad\quad\quad\quad \quad\quad\quad\quad \quad\quad\quad\quad \quad \sum_{v=1}^V q_{\boldsymbol{\psi}}(\boldsymbol{\eta} | \boldsymbol{x}_i,\boldsymbol{z}_i^{(v)})\big]\Big],
\end{split}
\label{eq:sivae_obj}
\end{equation}
as discussed in the next section.

\subsection{Optimization of LBD}\label{sec:optimization}
The optimization of (\ref{eq:obj}) with respect to global parameters $\boldsymbol{\theta}\setminus\boldsymbol{\alpha}$ can be performed by sampling a dropout mask for each data point in the mini-batch in the forward pass and calculating the gradients with respect to those parameters with SGD. The optimization with respect to dropout rates, however, is challenging as the reparameterization  technique (a.k.a. path-wise derivative estimator) \citep{kingma2013auto,rezende2014stochastic} cannot be used. On the other hand, score-function gradient estimators such as REINFORCE \citep{williams1992simple,fu2006gradient} possess high estimation variance.

In this paper, we estimate the gradient of (\ref{eq:obj}) with respect to $\boldsymbol{\alpha}$ with the ARM  estimator \citep{yin2018arm}.
Using ARM, we are able to directly optimize the Bernoulli dropout rates without introducing any bias. For a vector of $K$ binary random variables $\boldsymbol{z}=[z_1,\cdot \cdot \cdot, z_K]$ parameterized by $\boldsymbol{\alpha}=[\alpha_1,\cdot \cdot \cdot,\alpha_K]$, the logits of the Bernoulli probability parameters, the gradient of $\mathbb{E}_{\boldsymbol{z}}[\mathcal{L}(\boldsymbol{\theta},\boldsymbol{z})]$ with respect to $\boldsymbol{\alpha}$ can be expressed as \citep{yin2018arm}
\begin{equation}
\begin{split}
    \nabla_{\boldsymbol{\alpha}}\mathbb{E}_{\boldsymbol{z}}[\mathcal{L}(\boldsymbol{\theta},\boldsymbol{z})] =& \mathbb{E}_{\boldsymbol{u}\sim \prod_{k=1}^{K} \text{Unif}_{[0,1]}(u_k)} \Big[\big( \mathcal{L}(\boldsymbol{\theta},1_{[\boldsymbol{u}>\sigma(-\boldsymbol{\alpha})]})\\&-\mathcal{L}(\boldsymbol{\theta},1_{[\boldsymbol{u}<\sigma(\boldsymbol{\alpha})]})\big) \big(\boldsymbol{u} - \frac{1}{2} \big) \Big].
\end{split}
\end{equation}
Here $\text{Unif}$ denotes the uniform distribution, and
$\mathcal{L}(\boldsymbol{\theta},1_{[\boldsymbol{u}<\sigma(\boldsymbol{\alpha})]})$ denotes the loss obtained by setting 
$\boldsymbol{z}=1_{[\boldsymbol{u}<\sigma(\boldsymbol{\alpha})]}:=\big(1_{[u_1<\sigma(\alpha_1)]},\cdot \cdot \cdot, 1_{[u_K<\sigma(\alpha_K)]}\big)$.  With this gradient estimate, we can proceed to compute the gradient of our loss function in \eqref{eq:gen}, with special cases in \eqref{eq:Bayesian} and \eqref{eq:sivae_obj}. Note that the regularization term in \eqref{eq:gen} is usually not a function of $\boldsymbol{z}$. For example, the KL divergence term in many Bayesian and Bayesian deep learning formulations (e.g. the one in Section \ref{sec:Bayesian}) only depends on the distribution of the model parameters. With this, the gradient of the objective function with respect to the Bernoulli dropout parameters can be expressed as
 \begin{equation} \label{eq:armdrop}
 \begin{split}
  &\nabla_{\boldsymbol{\alpha}}\mathbb{E}_{\boldsymbol{z}}\big[\mathcal{L}(\boldsymbol{\theta},\boldsymbol{z} | \mathcal{D})\big] = \\ & \frac{N}{M} \sum_{i=1}^{M} \mathbb{E}_{\boldsymbol{u}_i\sim \prod_{k=1}^{K} \text{Unif}_{[0,1]}(u_{ik})} \Big[\big( \mathcal{E}
 (1_{[\boldsymbol{u}_i>\sigma(-\boldsymbol{\alpha})]})-\\&\quad \quad \mathcal{E}(1_{[\boldsymbol{u}_i<\sigma(\boldsymbol{\alpha})]})\big) \big(\boldsymbol{u}_i - \frac{1}{2} \big)  \Big]  + \nabla_{\boldsymbol{\alpha}}\mathcal{R}(\boldsymbol{\alpha)}.
  \end{split}
 \end{equation}
Here, $\mathcal{E}(1_{[\boldsymbol{u}_i>\sigma(-\boldsymbol{\alpha})]})$
is the empirical loss obtained by setting $\boldsymbol{z}$ to 1 if $\boldsymbol{u}_i>\sigma(-\boldsymbol{\alpha})$, and similarly for $\mathcal{E}(1_{[\boldsymbol{u}_i<\sigma(\boldsymbol{\alpha})]})$. Note that the expectation can be estimated using only one sample, allowing us to estimate the gradient efficiently.

\section{\uppercase{Results and Discussion}}

We evaluate LBD on three different tasks: image classification, semantic segmentation, and collaborative filtering. Before presenting these results, we investigate the performance on a toy example, where we can calculate the true gradients exactly, to demonstrate the advantages of LBD over the existing dropout schema. We implement all of our methods in Tensorflow \citep{Abadi:2016:TSL:3026877.3026899} on a single cluster node with Intel Xeon E5-2680 v4 2.40 GHz processor and a Tesla K80 Accelerator.

\subsection{Toy Example}

\begin{figure*}
\centering
\subfigure[LBD Bias]{
\includegraphics[width=0.22\textwidth]{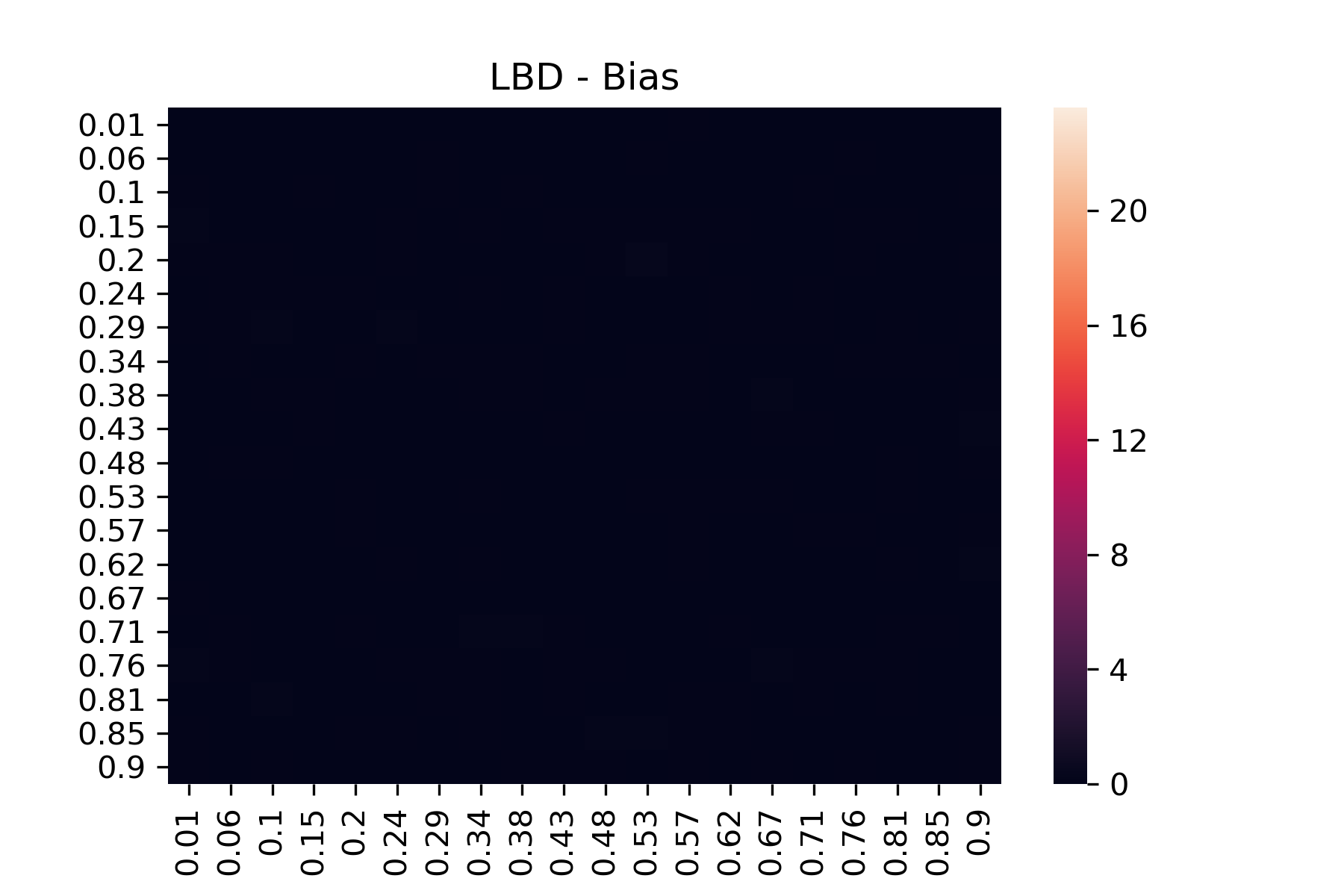}
\label{toy:1}
}
~
\subfigure[Concrete Bias]{
\includegraphics[width=0.22\textwidth]{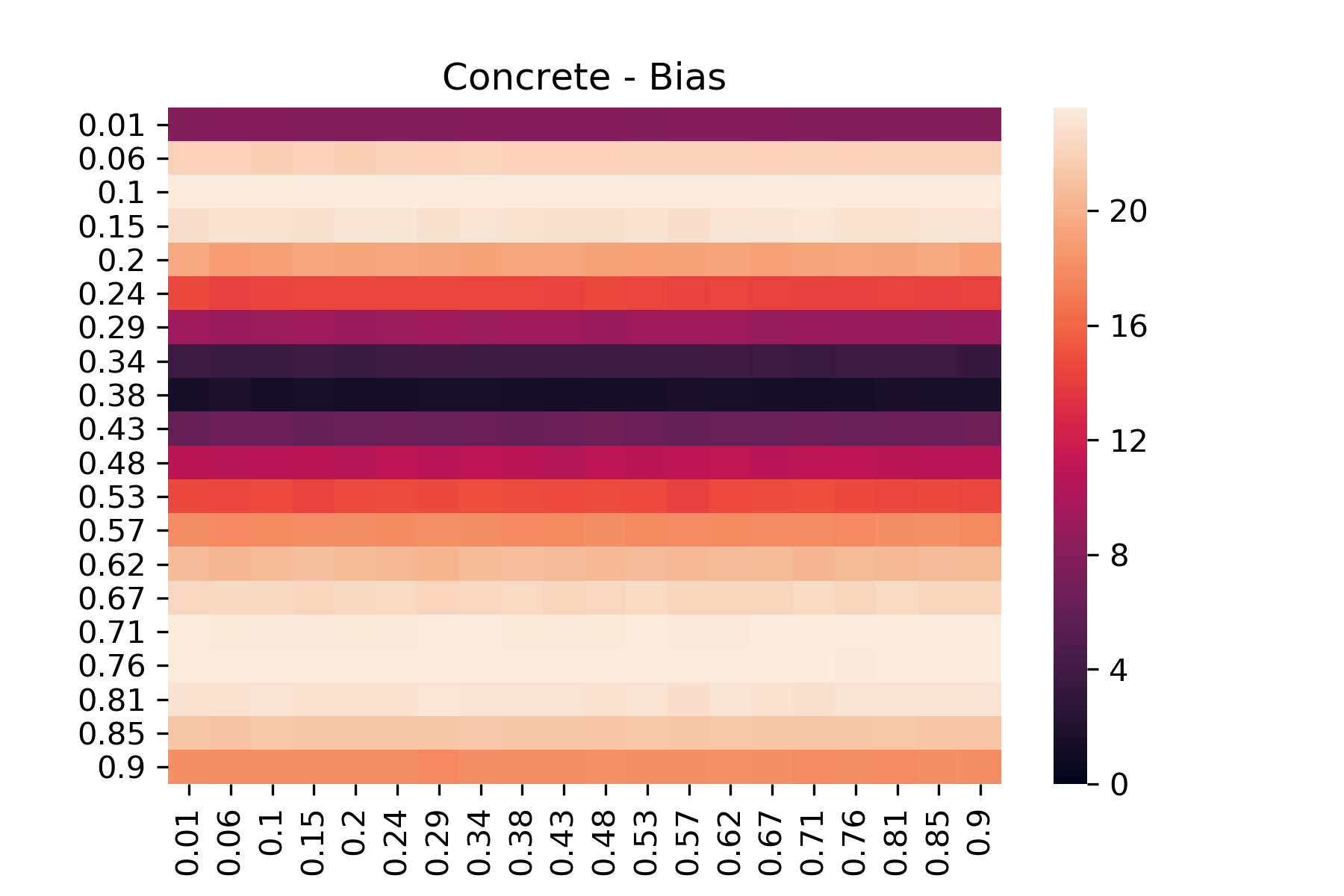}
\label{toy:2}
}
 ~
 \subfigure[LBD Std. Dev.]{
 \includegraphics[width=0.22\textwidth]{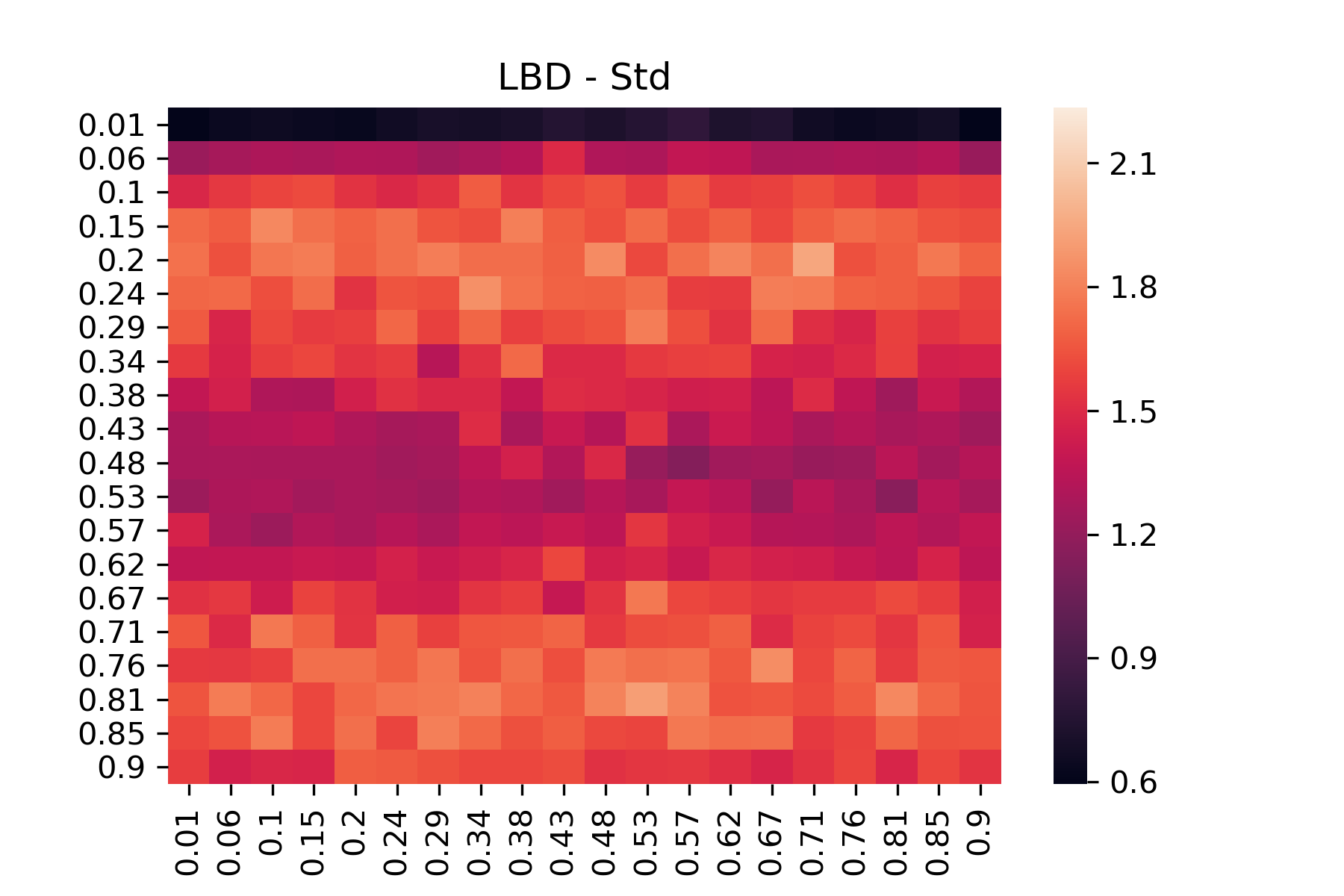}
 \label{toy:3}
 }
 ~
 \subfigure[Concrete Std. Dev.]{
 \includegraphics[width=0.22\textwidth]{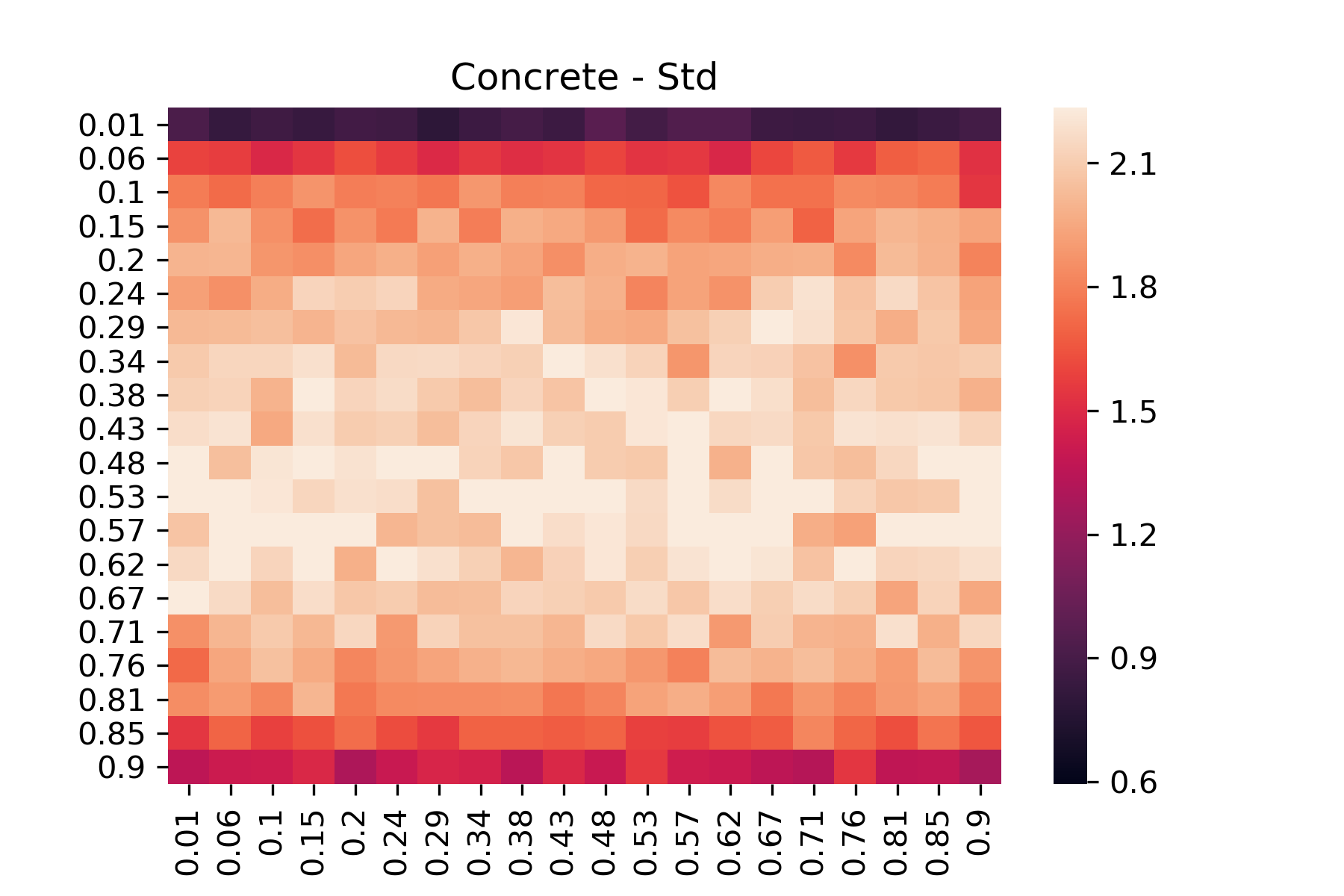}
 \label{toy:4}
 }
\\
\vspace{-0.12in}
\subfigure[LBD MSE]{
\includegraphics[width=0.22\textwidth]{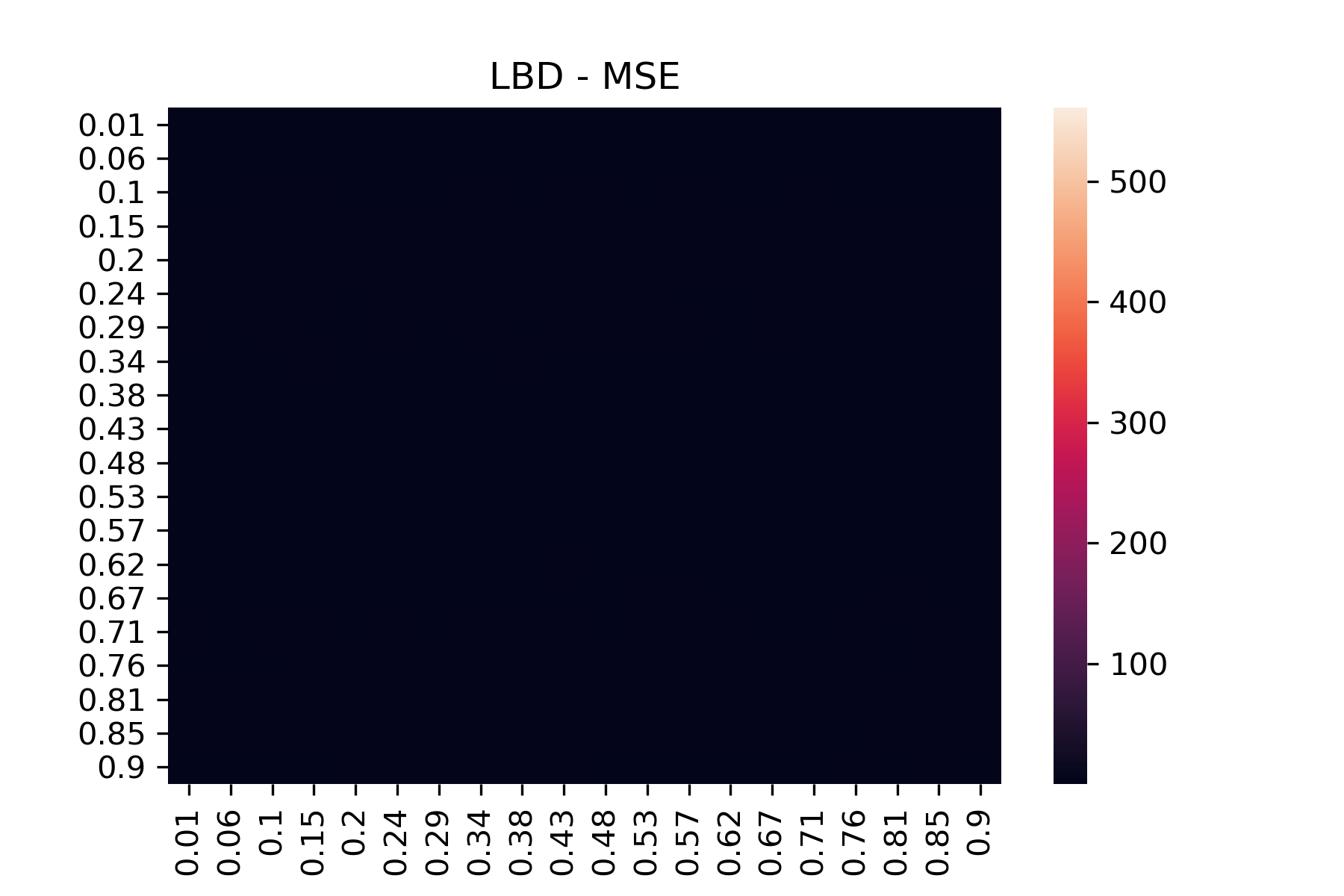}
\label{toy:5}
}
~
\subfigure[Concrete MSE]{
\includegraphics[width=0.22\textwidth]{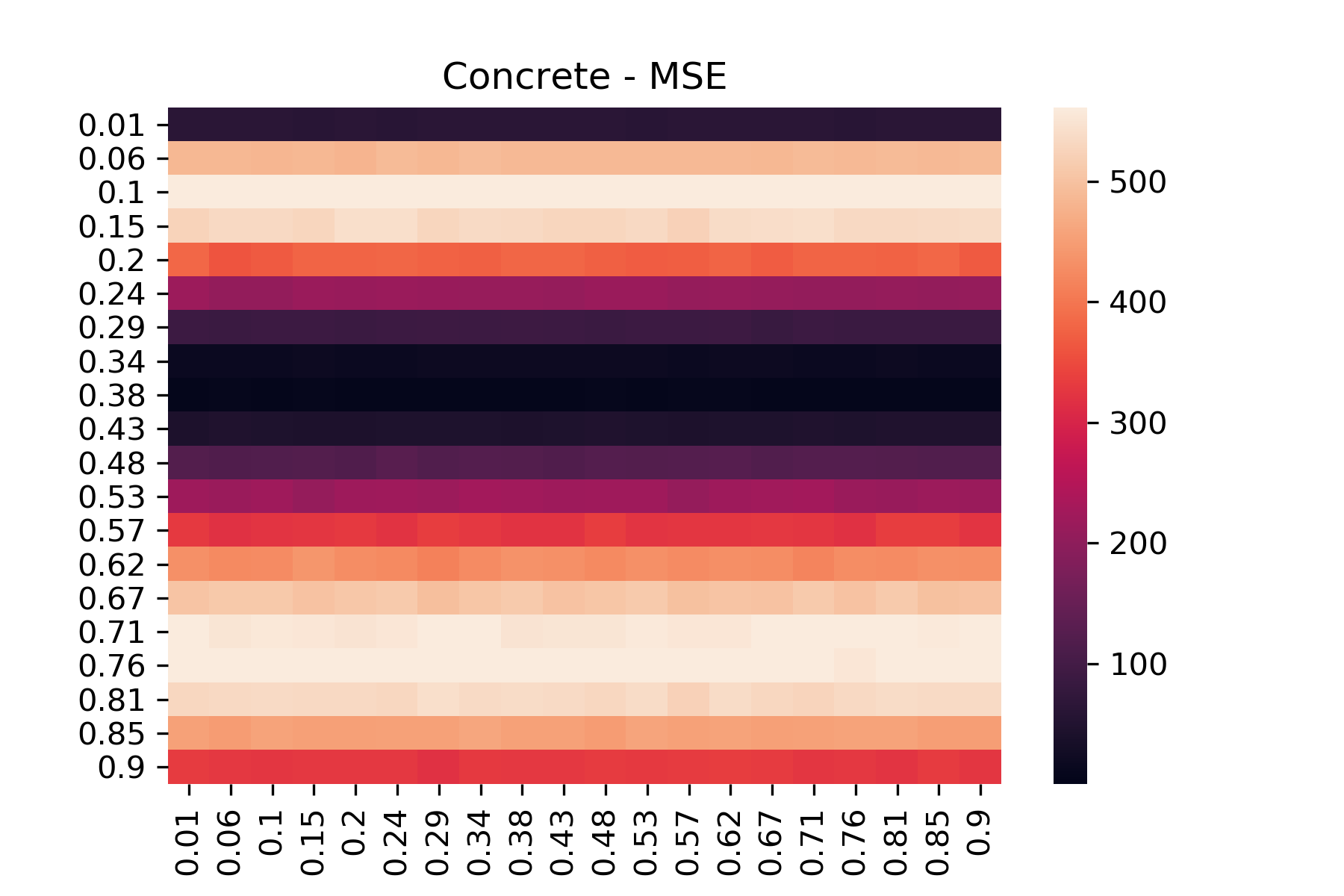}
\label{toy:6}
}
~
\subfigure[1st Param. Gradient]{
\includegraphics[width=0.22\textwidth]{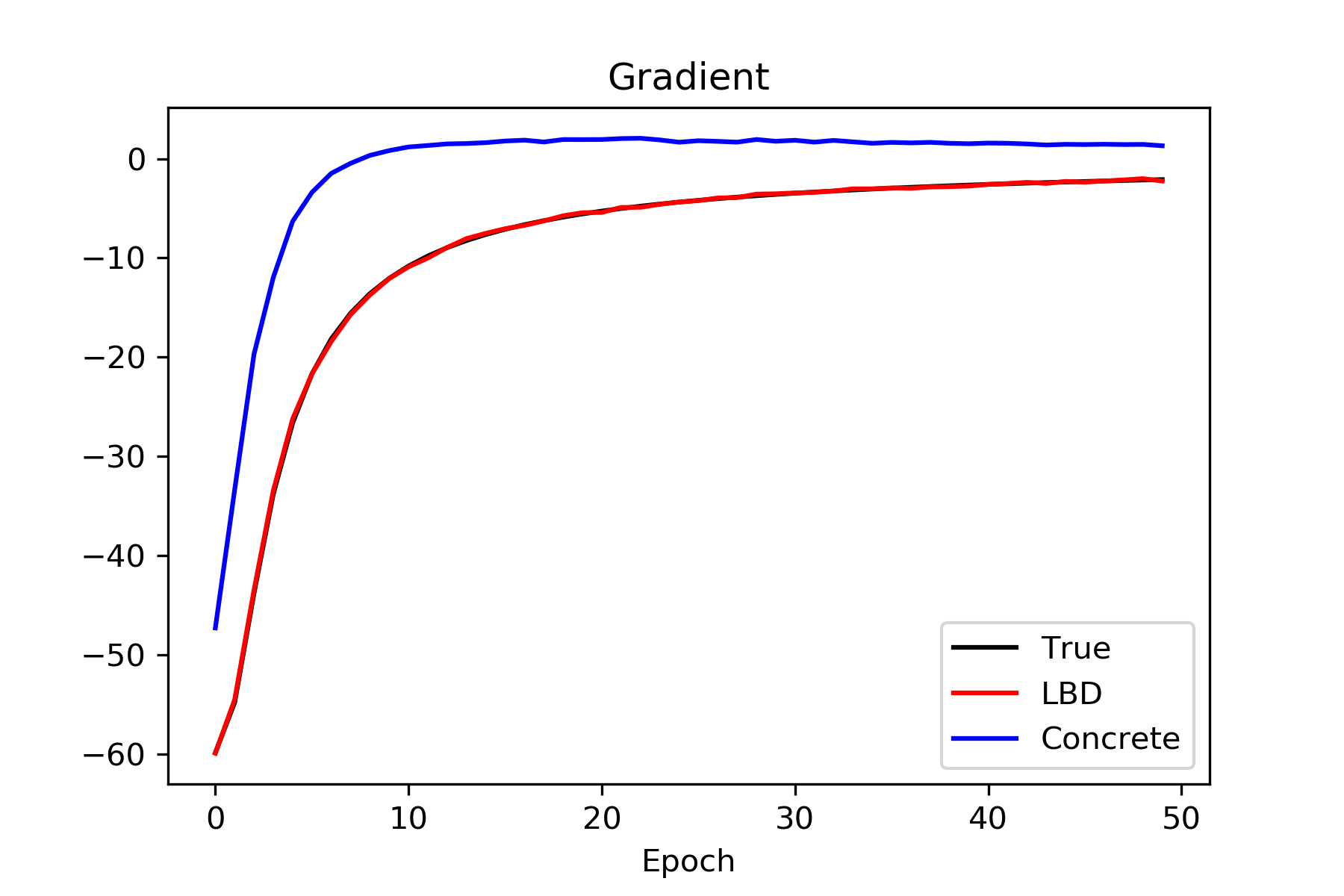}
\label{toy:7}
}
~
\subfigure[2nd Param. Gradient]{
\includegraphics[width=0.22\textwidth]{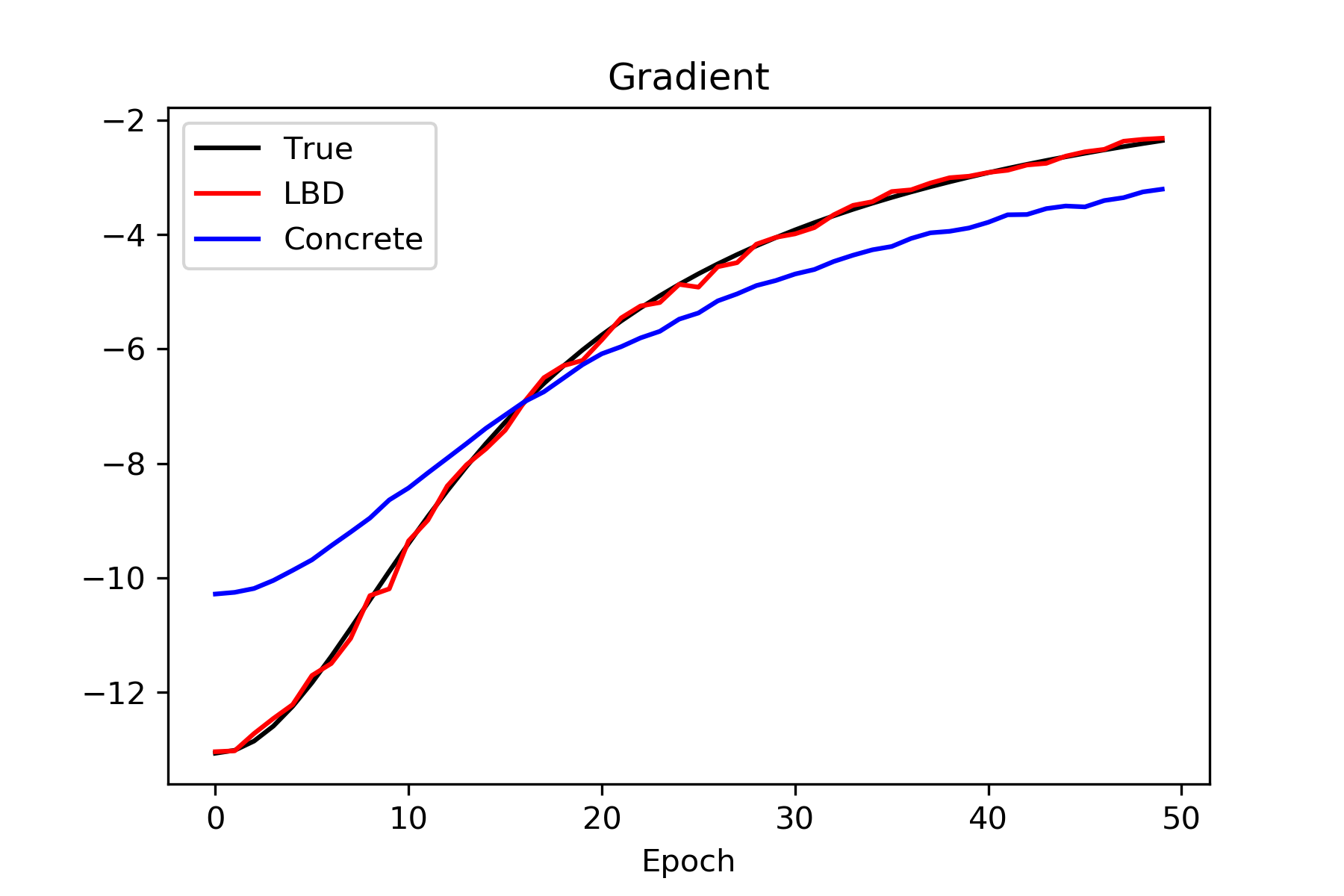}
\label{toy:8}
}
\caption{Comparison of gradient estimates for dropout parameters in the toy example.}
\vspace{-0.15in}
\label{fig:toy}
\end{figure*}

In this section, we investigate the bias and bias-variance trade-off of gradient estimates for dropout parameters from LBD and Concrete with respect to a simple Bernoulli dropout model where exact calculation of the gradients is tractable. For this purpose, we consider a toy regression task with simulated data. The base model for our example is a simple neural network with one input, one output and two hidden layer nodes, with ReLU non-linearity and dropout applied after hidden layer neurons. There are two dropout parameters, $\alpha_1$ and $\alpha_2$ for the hidden layer. The objective function is $\mathbb{E}_{\boldsymbol{z}_i\sim\prod_{i=1}^N \text{Ber}(\boldsymbol{z}_i;\sigma([\alpha_1,\alpha_2]))} \big[(y_i - f(x_i;W,z_i))^2\big]$. The data (3000 samples) is generated from the same model without the non-linearity and dropout (i.e. a linear model). The weights of the network are randomly initialized and held fixed during training. We calculate the true gradients for dropout parameters at different values of $\sigma(\alpha_1)$ and $\sigma(\alpha_2)$ and compare them with estimates by LBD and Concrete in the histograms of Figure \ref{fig:toy}. The bias, standard deviation~(STD), and mean squared error~(MSE) of the estimates for $\alpha_1$ from LBD and Concrete calculated by 200 Monte Carlo samples are shown in panels (a)-(f), respectively, while additional results for $\alpha_2$ are included in the Supplementary. LBD which leverages ARM clearly has no bias and lower MSE in estimating the gradients w.r.t. the Bernoulli model. We also provide trace plots of the estimated gradients by LBD and Concrete using 50 Monte Carlo samples when updating the parameters via gradient descent with the true gradients in the bottom panels in Figure \ref{fig:toy}. Panels (g) and (h) correspond to $\alpha_1$ and $\alpha_2$, respectively. The estimates from LBD follow the true gradients very closely.

\subsection{Image Classification and Semantic Segmentation}

In image classification and segmentation, we evaluate different dropout results based on prediction accuracy and mean Intersection over Union (IoU) respectively. To assess the quality of uncertainty estimation, we use the PAvPU ($P(\text{Accurate} | \text{Certain})$ vs $P(\text{Uncertain} | \text{Inaccurate})$) uncertainty evaluation metric proposed in \citep{mukhoti2018evaluating}. $P(\text{Accurate}  |  \text{Certain})$ is the probability that the model is accurate on its output given that it is confident on the same, while $P(\text{Uncertain}  |  \text{Inaccurate})$ is the probability that the model is uncertain about its output given that it has made a mistake in prediction. Combining these together results in the PAvPU metric, calculated by the following equation:
\begin{equation}
    \text{PAvPU} = \frac{n_{iu} + n_{ac}}{n_{ac}+n_{au}+n_{iu}+n_{ic}}.
\end{equation}
Here, $n_{ac}$ is the number of accurate and certain predictions; $n_{ic}$ is the number of inaccurate and certain predictions; $n_{au}$ is the number of accurate and uncertain predictions; and $n_{iu}$ is the number of inaccurate and uncertain predictions. Intuitively, this metric gives a high score to models that predict accurately with confidence or put high uncertainty estimates for incorrect predictions. To determine when the model is certain or uncertain, the mean predictive entropy can be used as a threshold. Or alternatively, different thresholds of the predictive entropy can be used: e.g. denoting the minimum and maximum predictive entropy for the validation dataset as $\text{PredEnt}_{\text{min}}$ and $\text{PredEnt}_{\text{max}}$, different thresholds $\text{PredEnt}_{\text{min}} + t(\text{PredEnt}_{\text{max}}-\text{PredEnt}_{\text{min}})$ can be considered by fixing $t\in [0,1]$. In other words, we would classify a prediction as uncertain if its predictive entropy is greater than a certain threshold.

\subsubsection{Image classification on CIFAR-10}

We evaluate LBD under different configurations on the CIFAR-10 classification task \citep{krizhevsky2009learning}, and compare its accuracy and uncertainty quantification with other dropout configurations on the VGG19 architecture \citep{simonyan2014very}. The CIFAR-10 dataset contains 50,000 training images and 10,000 testing images of size $32 \times 32$ from 10 different classes. The VGG architecture consists of two layers of dropout before the output. Each layer of dropout is preceded by a fully connected layer followed by a Rectified Linear Unit (ReLU) nonlinearity. We modify the existing dropout layers in this architecture using LBD as well as other forms of dropout. For all experiments, we utilize a batch size of 64 and train for 300 epochs using the Adam optimizer with a learning rate of $1\times10^{-5}$ \citep{kingma2014adam}. We use weights pretrained on ImageNet, and resize the images to $224 \times 224$ to fit the VGG19 architecture. No other data augmentation or preprocessing is done.

\begin{figure}[h]
\centering
\subfigure[CIFAR-10]{
\includegraphics[width=0.22\textwidth]{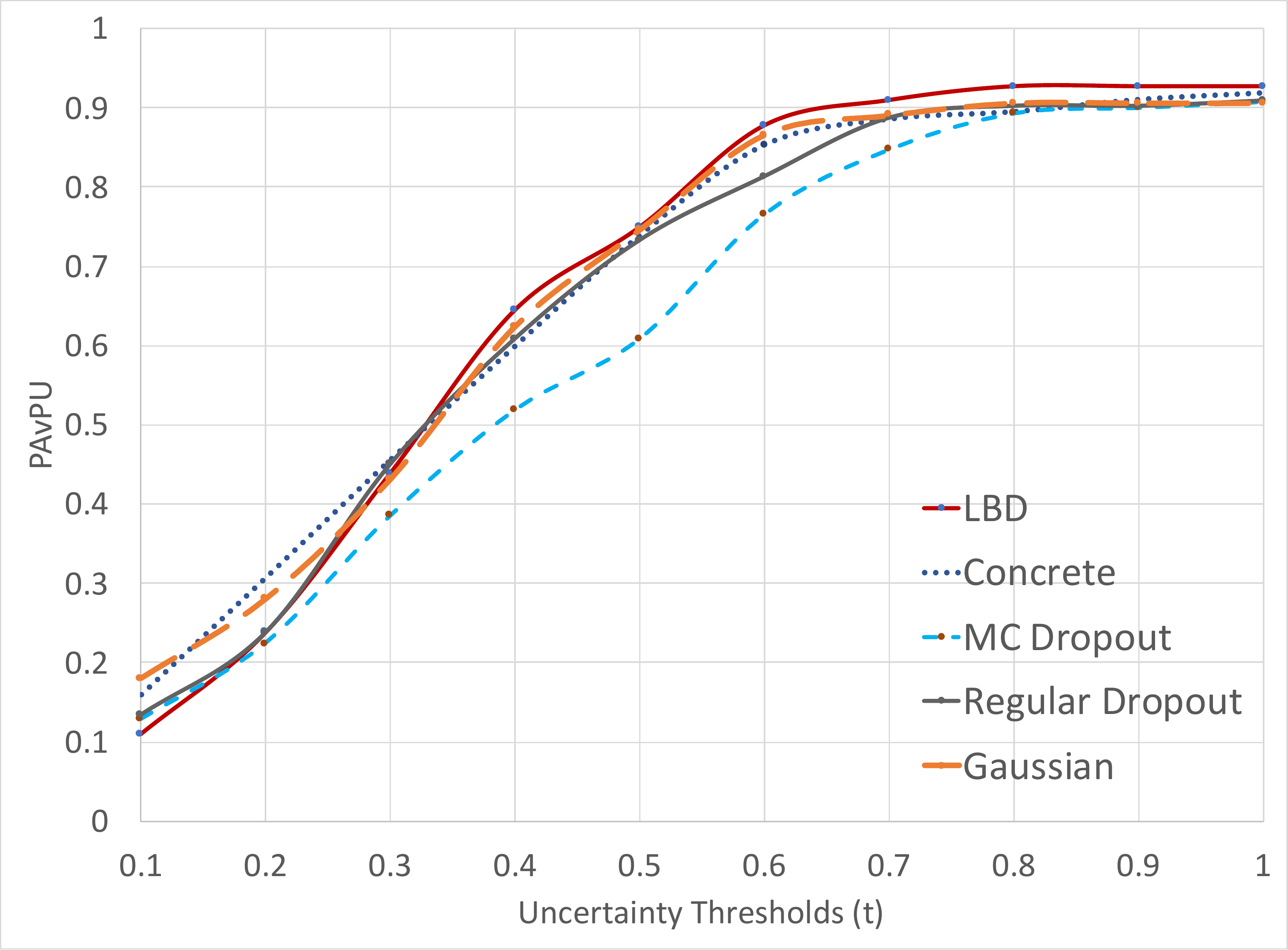}
\label{classification:pavpu}
}
\subfigure[CamVid]{
\includegraphics[width=0.22\textwidth]{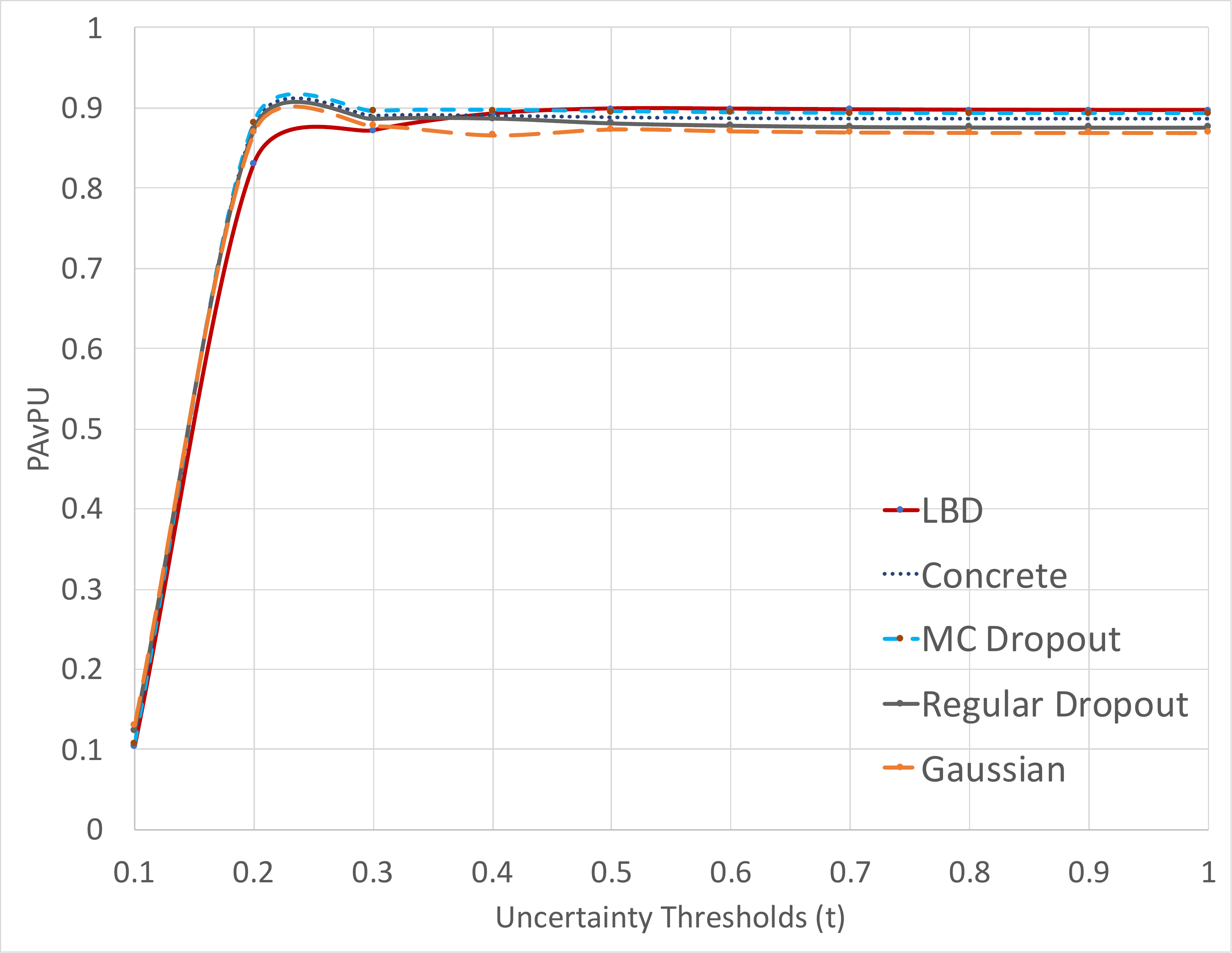}
\label{segmentation:pavpu}
}
\caption{PAvPU for different models under different uncertainty thresholds for classification on CIFAR-10 and segmentation for CamVid}
\vspace{-0.20in}
\end{figure}

We compare our LBD with regular dropout, MC dropout \citep{gal2016dropout}, and concrete dropout \citep{gal2017concrete}. For regular and MC dropout, we utilize the hand-tuned dropout rate of 0.5. To obtain predictions and predictive uncertainty estimates, we sample 10 forward passes of the architecture and calculate (posterior) mean and predictive entropy. The prediction accuracy and uncertainty quantification results for the last epoch of training is shown in Figure \ref{classification:pavpu} and Table \ref{classification:table}. Our LBD consistently achieves better accuracy and uncertainty estimates compared to other methods that learn or hand-tune the dropout rate. 

We further test LBD with a shared parameter for all dropout layers. The accuracy achieved in this case is 92.11\%, which is higher than the cases with hand-tuned dropout rates, but lower than LBD with different dropout rates for different layers included in Table \ref{classification:table}. This shows that even optimizing one dropout rate can still be beneficial compared to considering it as a hyperparameter and hand-tuning it. Moreover, it further confirms the need for the model adaptability achieved by learning more flexible dropout rates from data.

\begin{table*}
  \caption{Comparison of LBD accuracy and uncertainty quantification with other forms of dropout for classification on the CIFAR-10 dataset and semantic segmentation on the CamVid dataset. (All numbers in \%)}
  \centering
  \begin{tabular}{lll|lll} 
    \toprule
          & \multicolumn{2}{c}{CIFAR-10} & \multicolumn{3}{c}{CamVid} \\
    \midrule
    Method & Accuracy & Mean PAvPU & Accuracy & Mean IoU & Mean PAvPU \\
    \midrule
    LBD      & \bf{93.47} & \bf{64.52} & \bf{89.18} & \bf{60.42} & \bf{85.16} \\
    \hdashline
    Concrete & 92.72 & 59.97 & 88.64 & 55.95 & 83.81 \\
    Gaussian & 91.16 & 52.71 & 88.04 & 54.85 & 85.02 \\
    MC Dropout & 91.57 & 52.08 & 87.80 & 54.40 & 84.39 \\
    Regular Dropout & 91.61 & 56.81 & 86.65 & 53.62 & 81.17 \\
    \bottomrule
  \end{tabular}
  \label{classification:table}
  \vspace{-0.10in}
\end{table*}

\subsubsection{Semantic segmentation on CamVid}

We evaluate our LBD for image segmentation on the CamVid dataset \citep{brostow2009semantic}. This contains 367 training images, 101 validation images, and 232 testing images of size $480 \times 360$ with 11 semantic segmentation classes. In all experiments, we train on the training images and test on the testing images. We use the FC-DenseNet-103 architecture as the base model \citep{jegou2017one}. This architecture consists of 11 blocks, with the middle block having 15 layers. Each layer consists of a ReLU nonlinearity followed by a convolution and dropout. Due to the concatenation of many features within each resolution, dropout becomes important to regularize the network by removing less relevant features from consideration. In our experiments, we replace the dropout in the 15 middle layers of this architecture with LBD as well as other forms of dropout. Here, the dropout parameters are shared between neurons in the same layer. For all experiments, we utilize a batch size of 1 and train for 700 epochs. We did not crop the $480 \times 360$ sized image and performed horizontal image flipping for data augmentation. We train our model using the RMSProp optimizer using a decay of 0.995 \citep{hinton2012neural} and a learning rate of 0.0001. For regular and MC dropout, we utilize the hand-tuned dropout rate of 0.2.

For uncertainty estimates in image segmentation, each pixel can be individually classified into certain or uncertain; however, \citet{mukhoti2018evaluating} noted that capturing uncertainties occurring in regions comprising of multiple pixels in close neighborhoods would be more useful for downstream machine learning tasks. Thus, they opted to compute patch-wise uncertainties by averaging the entropy over a sliding window of size $w \times w$. In our evaluation, we use a window of size $2 \times 2$. The results of our semantic segmentation experiments are shown in Table \ref{classification:table} and Figure \ref{segmentation:pavpu}. As shown in the table and figure, our LBD consistently achieves better accuracy and uncertainty estimates, compared to other methods that either learn or hand-tune the dropout rates. It is also interesting to note that methods that are based on Bayesian approximation which learn the dropout parameters generally had better PAvPU than MC and regular dropout, confirming that Bayesian deep learning methods provide better uncertainty estimates when using optimized dropout rates.

\subsection{Collaborative Filtering for Implicit Feedback Data}

Collaborative filtering which predicts user preferences by discovering similarity patterns among users and items \citep{herlocker2004evaluating} is among the most notable algorithms for recommender systems. VAEs with multinomial likelihood have been shown to provide state-of-the-art performance for collaborative filtering
on implicit feedback data \citep{liang2018variational}. They are especially interesting for large-scale datasets due to the amortized inference.

Our experimental setup is similar to \citep{liang2018variational}. Following their heuristic search for $\beta$ in the VAE training objective, we also gradually increase $\beta$ from 0 to 1 during training and record the $\beta$ that maximizes the validation performance. For all variations of VAE and our SIVAE we use the multinomial likelihood. For all encoder and decoder networks the dimension of the latent representations is set to 200, with one hidden layer of 600 neurons. 

The experiments are performed on three user-item datasets: MovieLens-20M (ML-20M) \citep{harper2016movielens}, Netflix Prize (Netflix) \citep{bennett2007netflix}, and Million Song Dataset (MSD) \citep{bertin2011million}. We take similar pre-processing steps as in \citep{liang2018variational}. For ML-20M and Netflix, users who have watched less than 5 movies are discarded and user-movie matrix is binarized by keeping the ratings of 4 and higher. For MSD, users with at least 20 songs in their listening history
and songs that are listened to by at least 200 users are kept and the play counts are binarized. After pre-processing, the ML-20M dataset contains around 136,000 users and 20,000 movies with 10M interactions; Netflix contains around 463,000 user and 17,800 item with 56.9M interactions; MSD has around 571,000 user and 41,000 song with 33.6M interactions left.

 \begin{table*}
      \caption{Comparisons of various baselines and different configurations of VAE and dropout with multinomial likelihood on ML-20M, Netflix and MSD dataset. The standard errors are around 0.2\% for ML-20M and 0.1\% for Netflix and MSD. (All numbers in \%)}
  \label{cf-result}
  \centering
  
  \begin{tabular}{lccc}
     \toprule
     \multirow{2}{*}{Method} & \multicolumn{3}{c}{ML-20M/Netflix/MSD} \\
     \cline{2-4}
     & Recall@20 & Recall@50 & NDCG@100 \\
     \midrule
     SIVAE+LBDrop & \bf{39.95}/\bf{35.63}/\bf{27.18} & \bf{53.95}/\bf{44.70}/\bf{37.36}  & \bf{43.01}/\bf{39.04}/\bf{32.33}\\
     SIVAE+GDrop & 35.77/31.43/22.50 & 49.45/40.68/31.01  & 38.69/35.05/26.99 \\
     SIVAE+CDrop & 37.19/32.07/24.32 & 51.30/41.52/33.19  & 39.90/35.67/29.10 \\
     VAE-VampPrior+Drop & 39.65/35.15/26.26 & 53.63/44.43/35.89  & 42.56/38.68/31.37\\
     VAE-VampPrior & 35.84/31.09/22.02 & 50.27/40.81/30.33  & 38.57/34.86/26.57 \\
     VAE-IAF+Drop & 39.29/34.65/25.65 & 53.52/44.02/34.96  & 42.37/38.21/30.67 \\
     VAE-IAF &  35.94/32.77/21.92 & 50.28/41.95/30.23  & 38.85/36.23/26.50  \\
     VAE+Drop &  39.47/35.16/26.36 & 53.53/44.38/36.21 & 42.63/38.60/31.33 \\
     VAE  & 35.67/31.12/21.98 & 49.89/40.78/30.33 & 38.53/34.86/26.49 \\
     DAE & 38.58/34.50/25.89  & 52.28/43.41/35.23 & 41.92/37.85/31.04 \\
     \hdashline
     WMF & 36.00/31.60/21.10 & 49.80/40.40/31.20  & 38.60/35.10/25.70     \\
     SLIM & 37.00/34.70/- & 49.50/42.80/- & 40.10/37.90/-      \\
     CDAE & 39.10/34.30/18.80 & 52.30/42.80/28.30 & 41.80/37.60/23.70   \\
     \bottomrule
  \end{tabular}
  \vspace{-0.15in}
 \end{table*}

To evaluate the performance of different models, two of the most popular learning-to-rank scoring metrics are employed, Recall@$R$ and the truncated
normalized discounted cumulative gain (NDCG@$R$). Recall@$R$ treats
all items ranked within the first $R$ as equally important while NDCG@$R$ considers a monotonically increasing discount coefficient to emphasize the significance of higher ranks versus lower ones. Following \citep{liang2018variational}, we split the users into train/validation/test sets. All the interactions of users in training set are used during training. For validation and test sets, 80\% of interactions are randomly selected to learn the latent user-level representations, with the other 20\% held-out to calculate the evaluation metrics for model predictions. For ML-20M, Netflix, and MSD datasets 10,000, 40,000, and 50,000 users are assigned to each of the validation and test sets, respectively.
 
We compare the performance of VAE (with dropout at input layers) and DAE \citep{liang2018variational} (VAE+Drop and DAE) with our proposed SIVAE with learnable Bernoulli dropout (SIVAE+LBDrop). We also provide the results for SIVAE with learnable Concrete dropout (SIVAE+CDrop) and SIVAE with learnable Gaussian dropout (SIVAE+GDrop), where we have replaced the learnable Bernoulli dropout in our SIVAE construction of Section \ref{sec:SIVAE} with Concrete and Gaussian versions. Furthermore, the results for VAE without any dropout (VAE) are shown in Table \ref{cf-result}. Note that because of the discrete (binary) nature of the implicit feedback data, Bernoulli dropout seems a naturally better fit for the input layers compared with other relaxed distributions due to not changing the nature of the data. We also test two extensions of VAE-based models, VAE with variational mixture of posteriors (VAE-VampPrior) \citep{tomczak2018vae} and VAE with inverse autoregressive flow (VAE-IAF) \citep{kingma2016improved} with and without dropout layer for this task. The prior in VAE-VampPrior consists of mixture distributions with components given by variational posteriors conditioned on learnable pseudo-inputs, where the encoder parameters are shared between the prior and the variaional posterior. VAE-IAF uses invertible transformations based on an autoregressive neural network to build flexible variational posterior distributions. We train all models with Adam optimizer \citep{kingma2014adam}. The models are trained for 200 epochs on ML-20M, and for 100 epochs on Netflix and MSD. For comparison purposes, we have included the published results for weighted matrix factorization (WMF) \citep{hu2008collaborative}, SLIM \citep{ning2011slim}, and collaborative denoising autoencoder (CDAE) \citep{wu2016collaborative} on these datasets. WMF is a linear low-rank factorization model which is
trained by alternating least squares. SLIM is a sparse linear model which learns an item-to-item similarity matrix by solving a constrained $\ell_1$-regularization optimization. CDAE augments the standard DAE by adding per-user latent factors to the input.

From the table, we can see that the SIVAE with learnable Bernoulli dropout achieves the best performance in all metrics on all datasets. Note that dropout is crucial to VAE's good performance and removing it has a huge impact on its performance. Also, Concrete or Gaussian dropout which also attempt to optimize the dropout rates performed worse compared to VAE with regular dropout. 

\section{\uppercase{Conclusion}} 

In this paper we have proposed learnable Bernoulli dropout (LBD) for model-agnostic dropout in DNNs. Our LBD module improves the performance of regular dropout by learning adaptive and per-neuron dropout rates. We have formulated the problem for non-Bayesian and Bayesian feed-forward supervised frameworks as well as the unsupervised setup of variational auto-encoders. Although Bernoulli dropout is universally used by many architectures due to their ease of implementation and faster computation, previous approaches to automatically learning the dropout rates from data have all involved replacing the Bernoulli distribution with a continuous relaxation. Here, we optimize the Bernoulli dropout rates directly using the Augment-REINFORCE-Merge (ARM) algorithm. Our experiments on computer vision tasks demonstrate that for the same base models, adopting LBD results in better accuracy and uncertainty quantification compared with other dropout schemes. Moreover, we have shown that combining LBD with VAEs, naturally leads to a flexible semi-implicit variational inference framework with VAEs (SIVAE). By optimizing the dropout rates in SIVAE, we have achieved state-of-the-art performance in multiple collaborative filtering benchmarks. 

\bibliography{arxiv_version}

\end{document}